%% file: main.tex
\documentclass{applemlr}

\input{apple_preamble}

\usepackage{url}
\usepackage{pgfplots}
\pgfplotsset{compat=1.18}
\usepackage{float}

\title{RVPO: Risk-Sensitive Alignment via Variance Regularization}

\author{Ivan Montero}
\author{Tomasz Jurczyk}
\author{Bhuwan Dhingra}

\affiliation{Apple}

\abstract{
Current critic-less RLHF methods aggregate multi-objective rewards via an arithmetic mean, leaving them vulnerable to constraint neglect: strong performance in one objective (e.g., correctness) can offset critical failures in others (e.g., safety or formatting). This can mask low-performing ``bottleneck'' rewards that are necessary for reliable multi-objective alignment. We propose Reward-Variance Policy Optimization (RVPO), a risk-sensitive framework that penalizes variance across rewards during advantage aggregation, and shifts the objective from ``maximize sum'' to ``maximize consistency.'' Its SoftMin objective can be locally approximated by the mean reward minus an inter-objective variance penalty. We evaluate RVPO on rubric-based medical and scientific reasoning with up to 17 reward signals (Qwen2.5-3B/7B/14B) and on tool-calling with rule-based constraints (Qwen2.5-1.5B/3B).
RVPO improves overall scores on HealthBench (0.254 vs.\ 0.207 for GDPO at 14B, averaged across five training seeds). RVPO also maintains competitive accuracy on GPQA-Diamond while reducing late-stage degradation observed in other multi-reward methods.
}

\metadata[Correspondence]{\sffamily Ivan Montero: \url{ivamon@apple.com}; Tomasz Jurczyk: \url{tjurczyk@apple.com}; Bhuwan Dhingra: \url{bdhingra2@apple.com}}
\date{\sffamily\today}

\begin{document}

\maketitle

\section{Introduction}

Recent critic-less RLHF methods like Group Relative Policy Optimization (GRPO) \citep{shao2024deepseekmath} and Group Decoupled Policy Optimization (GDPO) \citep{liu2026gdpo} reduce memory overhead by removing the value network, but rely on the mean aggregation of rewards. Mean aggregation allows gains on one objective to offset failures on another, which we call \textbf{constraint neglect} (Figure~\ref{fig:aggregation_gap}).
Constrained optimization methods \citep{achiam2017constrained, dai2023safe} require explicit constraint thresholds, while the model-merging approach of \citet{rame2023rewarded} trains separate policies for different objectives. In our experiments, GDPO reaches mean best-checkpoint HealthBench scores of 43.1\% on Communication Quality and 12.3\% on Completeness at 7B across five training seeds.

To address constraint neglect, we introduce \textbf{Reward-Variance Policy Optimization (RVPO)}, a risk-sensitive framework that penalizes inter-reward variance.

\textbf{Our core contributions are as follows:}
\begin{enumerate}
    \item We identify and demonstrate the constraint neglect vulnerability inherent in mean-aggregated, critic-less multi-objective RL.
    \item We introduce RVPO, whose SoftMin objective can be locally approximated by the mean reward minus an inter-objective variance penalty. Its risk coefficient $k$ smoothly interpolates between mean and min aggregation.
    \item We validate RVPO across two multi-objective scenarios: LLM-judged rubric criteria (5--17 rewards), where RVPO improves HealthBench scores compared to GDPO across five seeds per scale and reduces late-training degradation, and rule-based tool-calling (2 rewards), where RVPO accelerates convergence on the bottleneck format constraint while preserving execution accuracy.
\end{enumerate}

\input{fig/aggregation_gap}

\section{Related Work}

\textbf{Reinforcement Learning from Human Feedback (RLHF) and Reward Hacking.} Large language model (LLM) alignment relies heavily on RLHF to steer models toward human preferences \citep{ouyang2022training, rafailov2023direct}. Current approaches are susceptible to reward hacking, where the policy may exploit underspecifications in the reward model to achieve high scores without improving quality \citep{chen2024accuracy, dai2025mitigating, gao2023scaling}. Length gaming is a common reward hacking example, where the model becomes exceedingly verbose. Recent work addresses this via length-normalized rewards, explicit penalties, or disentangled reward modeling \citep{meng2024simpo, park2024disentangling, chen2024odin}.

\textbf{Multi-Objective Alignment and Structured Evaluation.} Recent work decomposes rewards into structured rubric criteria \citep{gunjal2025rubrics, hashemi2024llm}, multi-attribute reward models \citep{wang2024helpsteer}, checklists \citep{viswanathan2025checklists}, and explicit safety rules \citep{mu2024rule}. A common approach to optimizing these decomposed signals is linear scalarization \citep{li2025optimizing, hayes2022practical}, which allows gains in one objective to compensate for failures in another. Other methods treat this as a constrained optimization problem requiring explicit constraint thresholds and separate cost models \citep{achiam2017constrained, paternain2019constrained, dai2023safe}, utilize post-hoc model merging \citep{rame2023rewarded}, or resolve gradient conflicts across objectives \citep{chen2026reward}. Inference-time approaches include steering frozen models via targeted intervention \citep{nguyen2025multi} or unified preference-aware reward models \citep{lin2025parm}.

\textbf{Critic-Less and Group-Relative Optimization.} Critic-less methods remove the Value Network by utilizing group-relative advantage estimation \citep{ahmadian2024back}, scaling successfully to emergent reasoning \citep{guo2025deepseek, yu2025dapo}. However, when extended to multi-objective settings, Group Relative Policy Optimization (GRPO) \citep{shao2024deepseekmath} suffers from scale dominance. Group Decoupled Policy Optimization (GDPO) \citep{liu2026gdpo} addresses this by normalizing individual reward models independently before summation. We build on GDPO; Section~\ref{sec:constraint_neglect} explains where its mean aggregation falls short.

\textbf{Risk-Sensitive and Variance-Penalized RL.} The inspiration of RVPO can be found in risk-sensitive Markov Decision Processes \citep{howard1972risk}, where agents optimize exponential-utility or variance-constrained objectives rather than expected returns \citep{hu2023tighter, zhong2025risk}. In particular, mellowmax \citep{asadi2017alternative} applies a LogSumExp soft-max over actions for smoother value estimation within a single-objective MDP. RVPO operates in a different setting: the soft-min is applied across concurrent reward channels within the advantage computation.
RVPO connects to risk-sensitive/exponential-utility objectives, isoelastic social-welfare aggregation \citep{atkinson1970measurement}, population-proportional preference aggregation \citep{kim2025beyond}, non-linear multi-objective scalarization \citep{roijers2013survey,hayes2022practical}, and maximin-style alignment \citep{chakraborty2024maxmin}.

\section{Background: Reward Aggregation and Constraint Neglect}
\label{sec:constraint_neglect}

Critic-less RLHF methods remove the Value Network by estimating advantages based on the intra-group relative performance of $G$ sampled responses. However, the aggregation method used determines the policy's vulnerability to constraint neglect.

\textbf{GRPO and Scale Dominance.}
Group Relative Policy Optimization (GRPO) \citep{shao2024deepseekmath} aggregates $M$ distinct objectives by summing the raw rewards for each generation, and then normalizing this total score across the group:

\[A_{GRPO} = \frac{\sum_{j=1}^{M} R_{j} - \mu_{total}}{\sigma_{total}},\]

where $R_j$ is the raw reward for the $j$-th objective, and $\mu_{total}$ and $\sigma_{total}$ are the mean and standard deviation of the summed rewards computed across the $G$ generations in the group.
Because raw rewards are summed directly, metrics with naturally large variances or unbounded scales (e.g., generation length or raw helpfulness) numerically dominate the advantage calculation. Small, sparse, or binary rewards (e.g., a strict penalty for a JSON schema violation) can be dominated by these quantities, making it hard for the model to learn rigid constraints.

\textbf{GDPO and Constraint Neglect.}
\label{sec:gdpo_constraint_neglect}
Group Decoupled Policy Optimization (GDPO) \citep{liu2026gdpo} addresses differing reward scales by normalizing the rewards independently. For each reward model $j$, GDPO computes a standard score ($Z_j$) across the $G$ generations, making all objectives have zero mean and unit variance. These $Z$-scores are then averaged:

\[A_{GDPO} = \frac{1}{M} \sum_{j=1}^{M} Z_{j} = \mu_{Z}.\]

GDPO addresses differing reward scales, but averaging still allows \textit{loss compensation}. A low score on a bottleneck constraint ($Z_{format} \ll 0$) can be offset by a high score on another objective ($Z_{length} \gg 0$).
Mean aggregation therefore allows compensation between training reward channels even after Z-normalization.

\section{Reward-Variance Policy Optimization}

RVPO replaces GDPO's mean with a soft minimum. For a given rollout $g$ from a group of $G$ generations, let $Z_j^{(g)} = (R_j^{(g)} - \mu_j) / \sigma_j$ denote the standardized reward on objective $j$, where $\mu_j$ and $\sigma_j$ are computed across the group. We use SoftMin aggregation:
\[A_{RVPO}^{(g)} = -\frac{1}{k} \ln\left(\frac{1}{M} \sum_{j=1}^{M} e^{-k Z_j^{(g)}}\right),\]
where $k > 0$ is the inverse temperature, which we term the \textbf{Risk Coefficient}. It moves from the mean to the minimum as $k$ grows:
\[
\lim_{k\to0} A_{RVPO}^{(g)} = \mu_Z^{(g)} = A_{GDPO}^{(g)},\qquad
\lim_{k\to+\infty} A_{RVPO}^{(g)} = \min_j Z_j^{(g)}.
\]
Here $\mu_Z^{(g)}=\frac{1}{M}\sum_j Z_j^{(g)}$ and $(\sigma_Z^{(g)})^2=\frac{1}{M}\sum_j(Z_j^{(g)}-\mu_Z^{(g)})^2$ are the mean and variance across objectives, not across rollouts. For small $|k(Z_j^{(g)}-\mu_Z^{(g)})|$, the local approximation is
\[A_{RVPO}^{(g)}\approx\mu_Z^{(g)}-\frac{k}{2}(\sigma_Z^{(g)})^2.\]
The detailed Taylor expansion is in Appendix~\ref{app:taylor}. We also evaluate the explicit variance-penalty alternative $A_{RVPO\text{-}explicit}^{(g)}=\mu_Z^{(g)}-\beta(\sigma_Z^{(g)})^2$, with $\beta>0$. Unlike this unbounded quadratic penalty, SoftMin approaches hard-min as $k$ increases.

\textbf{Power-mean / Nash connection.} Writing $\mathcal M_p(u)=(\frac{1}{M}\sum_j u_j^p)^{1/p}$ for the power mean of order $p\ne0$, the exponential change of variables $u_j=e^{Z_j}>0$ gives the identities
\begin{align*}
\log\mathcal M_p(e^Z)&=\frac{1}{p}\log\left(\frac{1}{M}\sum_j e^{pZ_j}\right),\\
\log\mathcal M_{-k}(e^Z)&=A_{RVPO}.
\end{align*}
The geometric/Nash limit is
\[
\mathcal M_0(u)=\lim_{p\to0}\mathcal M_p(u)=\left(\prod_j u_j\right)^{1/M},\qquad
\log\mathcal M_0(e^Z)=\mu_Z=A_{GDPO}.
\]
Thus GDPO and RVPO lie on one interpolation family: $p\to0$ gives the geometric/Nash point on $u$, $p=-k<0$ gives RVPO, and $p\to-\infty$ gives hard-min. The exponential transformation is necessary because standardized rewards can be signed.

The sensitivity of the SoftMin advantage to each normalized reward is $\partial A_{RVPO}/\partial Z_j = w_j=e^{-kZ_j}/\sum_\ell e^{-kZ_\ell}$. For each generation, channels with lower normalized reward scores receive more weight, and that extra weight decreases as their scores improve relative to the other channels. Finite $k$ retains contributions from several weak channels rather than assigning all pressure to the minimum. This bottleneck-aware feedback makes compensation less desirable. The full procedure is summarized in Algorithm~\ref{alg:rvpo} (Appendix).

SoftMin is always at most the arithmetic mean across channels, and the gap grows with inter-objective disagreement; this is the intended penalty on uneven generations.

Aggregating raw reward components before group normalization is one option, as in GRPO-explicit. We normalize first because channels differ in scale and sparsity, making them comparable before SoftMin determines their relative influence.

\section{Experimental Setup}

We compare RVPO with GRPO \citep{shao2024deepseekmath} and GDPO \citep{liu2026gdpo} on rubric-based reasoning and rule-based tool calling.

\subsection{Environments and Reward Formulation}
\textbf{Rubrics-as-Rewards (LLM-Judged Constraints):} We evaluate RVPO using the Rubrics-as-Rewards (RaR) framework \citep{gunjal2025rubrics}, where the number of reward signals varies dynamically per prompt (5--17 criteria). We evaluate on two domains from \citet{gunjal2025rubrics}: RaR-Medicine (20k clinical reasoning prompts) and RaR-Science (20k graduate-level science prompts). Unlike prior RaR implementations that collapse evaluations into a single scalar, we treat each of the $M$ criteria as an independent reward channel.
For each criterion, \texttt{gpt-4o-mini} acts as the judge, outputting a binary satisfaction score. The RaR framework assigns priority weights to each criterion, which we incorporate pre-normalization; we evaluate post-normalization weighting in Appendix~\ref{app:weighted_rewards}.

\textbf{Tool Calling (Rule-Based Constraints):} We use RLLA-4k \citep{qian2025toolrl}, a curated subset of 4,000 tool-calling trajectories, to evaluate multi-step reasoning alongside structural constraints. We define two competing reward signals: Execution Correctness, a continuous scalar evaluating tool and parameter matching against the ground truth, and Format Adherence, a binary constraint verifying XML schema compliance.

\subsection{Training Implementation and Baselines}

All experiments were run on a single node with 8 NVIDIA H100 GPUs. We train Qwen2.5 models \citep{yang2024qwen2} using the \texttt{verl} \citep{sheng2024hybridflow} and \texttt{TRL} frameworks: 1.5B and 3B for tool-calling; 3B, 7B, and 14B for rubrics-as-rewards. For rubrics, $k$ and $\beta$ were selected separately for Medicine and Science at 7B, favoring final-checkpoint performance over the highest early peak, then transferred unchanged to 3B and 14B. Medicine uses $k=0.5\to2.0$, $\beta=1.0$; Science uses $k=1.0\to2.0$, $\beta=0.5$. This was in-domain selection, without a separate held-out development set. Full training hyperparameters are in Appendix \ref{app:hyperparameters}.

Across domains, we compare RVPO against standard GRPO (which sums raw rewards) and GDPO (which independently Z-normalizes channels before summing). For rubric-based RVPO, the risk coefficient $k$ is linearly annealed from $k_{\text{start}}$ to $k_{\text{end}}$ (denoted $k = k_{\text{start}} \to k_{\text{end}}$). This shifts aggregation from near-mean aggregation towards greater emphasis on bottleneck channels. In the rubrics setting, we include two single-reward baselines from the RaR framework \citep{gunjal2025rubrics}: GRPO (Explicit), which aggregates per-criterion scores into a single weighted scalar; and GRPO (Implicit), the strongest baseline in the original work, where the LLM judge outputs a single holistic score.

\input{tab/rubrics_results_scale} 

\subsection{Evaluation Methodology}

For models trained on the RLLA-4k tool-calling dataset, we evaluate using the AST-based metrics of the Berkeley Function Call Leaderboard v3 (BFCL-v3) \citep{patil2025bfcl}, averaging results across five independent runs per method. For models trained on RaR-Medicine, we evaluate multi-objective alignment using the full 5,000-example HealthBench framework \citep{arora2025healthbench}, whose rubric scoring was validated against physician preferences. HealthBench explicitly scores models across five independent rubric axes: \textit{Communication Quality, Instruction Following, Accuracy, Context Awareness,} and \textit{Completeness}. Finally, for models trained on RaR-Science, we evaluate loglikelihood accuracy on the GPQA-Diamond benchmark \citep{rein2024gpqa}. For Medicine, we report mean and standard deviation over five distinct training seeds per method and scale. Checkpoints are evaluated every 50 steps over 300 steps (six evaluations); best is selected separately within each run and final is step 300. Science results remain single-run and are reported separately in Appendix~\ref{app:science}. GPQA-Diamond and BFCL-v3 serve as capability parity checks in the evaluated settings.

\section{Results}

\subsection{Rubrics-as-Rewards}
\label{sec:rar_medicine}


Table~\ref{tab:rubrics_results} reports the five-seed HealthBench comparison. RVPO has higher mean best- and final-checkpoint scores than GDPO at all three scales, reaching a mean best score of $0.254\pm0.017$ versus $0.207\pm0.032$ at 14B. The clearest pattern is improved end-of-training retention: final scores improve over GDPO by $0.043/0.130/0.175$ at 3B/7B/14B. RVPO also has higher mean scores than GRPO-explicit at both checkpoints across scales. At 7B, GRPO-implicit and RVPO-explicit remain competitive, with mean final scores of $0.160$ versus RVPO's $0.152$. Across repeated runs, RVPO reduces late-training degradation while remaining competitive at the best checkpoint.

Training diagnostics show persistent inter-rubric disagreement under GDPO: the standard deviation across active rubric Z-scores, averaged over generations and training, is $0.635\pm0.003 / 0.599\pm0.015 / 0.602\pm0.006$ at 3B/7B/14B (mean $\pm$ seed standard deviation). At the best checkpoint, Completeness is the weakest mean HealthBench axis at all scales, with Communication Quality $4.0/3.5/2.9\times$ higher. These summaries show persistent disagreement during training and imbalanced downstream performance; they do not establish a sample-level correlation between the two.

\input{fig/healthbench_axes}

Figure~\ref{fig:healthbench_best} illustrates the per-axis trade-off in single-run 7B experiments. In these runs, GDPO scores highly on Communication Quality but poorly on Completeness and Context Awareness. RVPO improves Accuracy, Context Awareness, and Completeness while giving up some Communication Quality and Instruction Following performance. This illustrates how bottleneck improvements can accompany higher overall performance; Table~\ref{tab:rubrics_results} provides the five-seed overall comparison.

GPQA-Diamond evaluates whether balanced rubric optimization generalizes to a held-out reasoning benchmark. While RaR-Science covers graduate-level topics, GPQA tests specialized expert knowledge requiring different reasoning patterns. The single-run results on this small benchmark ($N=198$; Appendix~\ref{app:science}) show broadly comparable accuracy, providing a parity check.

\subsection{Tool Calling Dynamics}

\input{tab/bfclv3}

\input{fig/tool_plots}

In the tool-calling setting, execution correctness is a continuous signal and format adherence is a sparse, binary constraint.
As shown in Figure \ref{fig:tool_calling_results}, GRPO is highly sensitive to this disparity: the larger magnitude of the correctness reward overshadows the binary format reward, delaying formatting improvements until step 25. GDPO mitigates this scale disparity by normalizing the reward distributions ($Z_j$), leading to earlier formatting improvements.

GDPO plateaus before reaching full format compliance. This is consistent with compensation between the two reward channels. RVPO improves convergence on format adherence while reducing the run-to-run variation observed in the baselines.

Table~\ref{tab:bfcl_results} shows that all methods achieve comparable downstream accuracy on BFCL-v3 across 1.5B and 3B scales. This indicates that RVPO's primary benefit in this low-dimensional setting ($M{=}2$) is faster convergence on the bottleneck constraint during training.

\subsection{Risk Coefficient Selection}
\label{sec:ablation_k_sweep}

The 7B Medicine selection favored the annealed $k=0.5\to2.0$ schedule over the higher-peak but less stable static $k=5.0$ setting. Appendix~\ref{app:k_sweep} reports the single-run selection sweep and gives a training-statistics rule of thumb for the scale of $k$. The selected settings transfer unchanged to 3B and 14B; tool-calling uses static $k=1.0$. We retain RVPO-explicit as an ablation; the single-run $\beta$ sweep in Appendix~\ref{app:beta_sweep} illustrates its sensitivity to the penalty coefficient.

\subsection{Controlled Corruption}

We test a fixed weak rubric channel on 3B RaR-Medicine with five seeds per cell. Under symmetric label noise at $p=0.5$, the target receives $1.194\pm0.010$ times equal-share aggregation weight under RVPO, versus $2.095\pm0.042$ under hard-min. RVPO assigns less weight to the corrupted channel than hard-min. Post-normalization weighting can separately encode a declared priority. Appendix~\ref{app:corruption} reports the complete conditions and results.

\section{Limitations and Future Work}

The primary limitation of RVPO is the sensitivity of the risk coefficient $k$. The choice of $k$ remains sensitive to reward space dimensionality, group size $G$ (our experiments use $G{=}4$ for tool-calling and $G{=}16$ for rubrics), and inter-objective conflict, motivating future work on adaptive scheduling. The settings selected at 7B transfer across scales, but broader validation of the training-statistics heuristic remains open (Appendix~\ref{app:k_sweep}). Additionally, as demonstrated in Appendix \ref{app:weighted_rewards}, RVPO's bottleneck prioritization is driven by per-channel normalized scores rather than declared priority; post-normalization weighting can encode priorities (Appendix~\ref{app:corruption}). RVPO may also amplify noise from unreliable reward channels, as the soft-min focuses optimization on whichever objective produces the lowest Z-scores regardless of whether this reflects policy weakness or reward model noise. Our corruption tests cover independent per-generation errors at 3B; Appendix~\ref{app:corruption} details their scope and downstream effects. Constrained-optimization and Pareto-style comparisons remain useful future directions.

\section{Conclusion}

We introduced Reward-Variance Policy Optimization (RVPO), which addresses constraint neglect in mean-aggregated multi-objective RL via a LogSumExp variance penalty. RVPO scales from two rule-based rewards to 17 concurrent LLM-judged criteria across multiple model sizes, improving mean best- and final-checkpoint HealthBench scores over GDPO across five seeds per scale while reducing late-training degradation.

\subsubsection*{Acknowledgments}
We thank Xinyan Velocity Yu, Eric Peng, Nitin Gupta, Deepika Bablani, Russ Webb, and Dong Yin for feedback on early drafts of this work.

\bibliographystyle{iclr2027_conference}
\bibliography{references}

\clearpage
\appendix

\section{Appendix}

\subsection{RVPO Algorithm}

Algorithm~\ref{alg:rvpo} summarizes the full RVPO training procedure, including per-channel Z-normalization, inactive reward masking, and SoftMin aggregation. In the RaR setting, the number of active reward channels varies per prompt ($M_{\text{active}} \in [5, 17]$), as each prompt defines a distinct rubric; channels corresponding to criteria not applicable to the current prompt are excluded from the aggregation. The risk coefficient $k(t)$ may be held constant or annealed (e.g., linearly) over training steps. As $k \to 0$, RVPO recovers standard GDPO (mean aggregation); as $k \to \infty$, it reduces to hard-min.

\begin{algorithm}[h]
\caption{Reward-Variance Policy Optimization (RVPO).}
\label{alg:rvpo}
\begin{algorithmic}[1]
\REQUIRE Policy $\pi_\theta$, reference policy $\pi_{\text{ref}}$, $M$ reward functions $\{R_j\}_{j=1}^M$, group size $G$, risk coefficient schedule $k(t)$
\FOR{each training step $t$}
    \STATE Sample batch of prompts $\{x_i\}$
    \FOR{each prompt $x_i$}
        \STATE Generate $G$ responses: $\{y_i^{(g)}\}_{g=1}^G \sim \pi_\theta(\cdot \mid x_i)$
        \STATE Compute rewards: $R_j^{(g)} = R_j(x_i, y_i^{(g)})$ for $j = 1, \ldots, M$
        \FOR{each reward channel $j = 1, \ldots, M$}
            \STATE $\mu_j = \frac{1}{G}\sum_{g} R_j^{(g)}$, \quad $\sigma_j = \text{std}(\{R_j^{(g)}\}_g)$ \COMMENT{Z-normalize}
            \STATE $Z_j^{(g)} = (R_j^{(g)} - \mu_j) / (\sigma_j + \epsilon)$
        \ENDFOR
        \STATE $A^{(g)} = -\frac{1}{k(t)} \ln\!\left(\frac{1}{M_{\text{active}}} \sum_{j \in \text{active}} e^{-k(t) \cdot Z_j^{(g)}}\right)$ \COMMENT{SoftMin over active channels}
    \ENDFOR
    \STATE Whiten advantages across batch: $\hat{A}^{(g)} = (A^{(g)} - \mu_A) / \sigma_A$
    \STATE Update $\theta$ via clipped policy gradient with advantages $\hat{A}^{(g)}$ and KL penalty against $\pi_{\text{ref}}$
\ENDFOR
\STATE \textbf{Note:} For RVPO-explicit, replace the SoftMin aggregation with $A^{(g)} = \mu_Z^{(g)} - \beta \cdot \left(\sigma_Z^{(g)}\right)^2$ for active channels only.
\end{algorithmic}
\end{algorithm}

\subsection{Training Hyperparameters}
\label{app:hyperparameters}

This section details the hyperparameter configurations used for the experiments. All training was conducted on a single node with 8 NVIDIA H100 GPUs. On this hardware, tool-calling training runs require $\approx$1.5 hours for 1.5B models and $\approx$2.5 hours for 3B models. In the rubrics setting, 300-step training runs require $\approx$2 hours (3B), $\approx$4.5 hours (7B), and $\approx$8 hours (14B) per experiment. RVPO's aggregation step (LogSumExp over $M$ channels) adds negligible wall-clock overhead ($<$1\%) relative to GDPO.

\textbf{Tool Calling (RLLA-4k):} To ensure a rigorous baseline comparison, we utilize the exact training recipe established by GDPO \citep{liu2026gdpo} and ToolRL \citep{qian2025toolrl}. We use the \texttt{verl} framework \citep{sheng2024hybridflow} to finetune Qwen2.5-Instruct models (1.5B and 3B) for 15 epochs ($\approx$117 steps) with a PPO mini-batch size of 128, a learning rate of $1 \times 10^{-6}$, a group size of $G=4$ rollouts per prompt, and an adaptive KL penalty ($\beta_{KL}=0.001$). We use a fixed risk coefficient $k=1.0$.

\textbf{Rubrics-as-Rewards (RaR-Medicine and RaR-Science):} Similarly, we adopt the training hyperparameters established by the Rubrics-as-Rewards framework \citep{gunjal2025rubrics}. We use the \texttt{TRL} framework to finetune Qwen2.5 base policies (3B, 7B, 14B) for 300 steps with an effective batch size of 96 prompts per step. We apply a learning rate of $5 \times 10^{-6}$ with a 10\% linear warmup schedule, sampling $G=16$ responses per prompt at a temperature of 1.0, and a KL penalty coefficient of $\beta_{KL}=0.04$ against the frozen reference policy. Checkpoints are saved and evaluated every 50 steps; we report the best-performing checkpoint alongside the final checkpoint (step 300).

\textbf{Asset Licenses:} Qwen2.5 models (Apache 2.0), RLLA-4k (Apache 2.0), RaR-Medicine/Science (CC-BY 4.0), HealthBench (MIT), GPQA-Diamond (CC-BY 4.0), BFCL-v3 (Apache 2.0), TRL (Apache 2.0), verl (Apache 2.0).

\subsection{Science Evaluation}
\label{app:science}

Table~\ref{tab:science_results} reports the GPQA-Diamond results, with one training run per configuration and separate best and final checkpoints.
\input{tab/science_results}

\subsection{Local Mean--Variance Approximation}
\label{app:taylor}

To build intuition for the local mean--variance interpretation of SoftMin, we perform a second-order Taylor expansion around the mean $\mu_Z^{(g)}$. Let $Z_j^{(g)} = \mu_Z^{(g)} + \delta_j$, such that the mean deviation $\frac{1}{M}\sum \delta_j = 0$ and the variance $\frac{1}{M}\sum \delta_j^2 = \left(\sigma_Z^{(g)}\right)^2$. By factoring out $e^{-k\mu_Z^{(g)}}$ and applying the approximations $e^y \approx 1 + y + \frac{y^2}{2}$ (for small $y$) alongside $\ln(1+y) \approx y$, the objective simplifies as follows:

\begin{align*}
    A_{RVPO}^{(g)} &= \mu_Z^{(g)} - \frac{1}{k} \ln\left(\frac{1}{M} \sum_{j=1}^{M} e^{-k\delta_j}\right) \\
    &\approx \mu_Z^{(g)} - \frac{1}{k} \ln\left(1 + \frac{k^2}{2}\left(\sigma_Z^{(g)}\right)^2\right) \\
    &\approx \mu_Z^{(g)} - \frac{k}{2}\left(\sigma_Z^{(g)}\right)^2 = A_{RVPO\text{-}explicit}^{(g)}(k/2).
\end{align*}

This is a local approximation for small $|k\delta_j|$, giving an effective variance coefficient $\beta=k/2$ in that regime.

\subsection{Hyperparameter Selection Details}
\label{app:k_sweep}

Table~\ref{tab:k_sweep_appendix} reports the single-run 7B Medicine selection sweep. Settings were selected separately for Medicine and Science, favoring final-checkpoint performance, and then transferred unchanged to 3B and 14B. The selected explicit-variance coefficients are $\beta=1.0$ for Medicine and $\beta=0.5$ for Science. Table~\ref{tab:rubrics_results} reports five-seed results for the selected settings.
\input{tab/k_sweep_ablation}

Beyond the selection protocol, equating the small-$k$ penalty scale $\frac{k}{2}(\sigma_Z^{(g)})^2$ with the worst-channel scale $|\min_j Z_j^{(g)}|$ gives the rule of thumb
\[
k^*=\frac{2\,\mathbb E_g[|\min_j Z_j^{(g)}|]}{\mathbb E_g[(\sigma_Z^{(g)})^2]}.
\]
In the analyzed rubric runs, it converged near $4.4$ within the first few steps and remained there. The Medicine schedule ends below this crossover at $2.0$, where SoftMin still combines information from several channels. This gives a rule of thumb for choosing the scale of $k$ from a short pilot or the opening training steps; it is not an exact optimum or a separately validated selection protocol. Validation across additional group sizes, channel counts, and conflict levels remains future work.

\subsection{Pre- vs.\ Post-Normalization Weighting}
\label{app:weighted_rewards}

Table~\ref{tab:weighted_ablations} evaluates methods under a weighted regime on the HealthBench (Medicine) domain. Standard RVPO incorporates the AI-generated categorical priority weights from the RaR framework \citep{gunjal2025rubrics} (e.g., Essential=1.0, Optional=0.3) directly into the raw reward computation ($R_j = c_j \cdot w_j$, where $c_j \in \{0,1\}$ is the binary criterion score). However, for binary criteria, these scalar weights are naturally absorbed by the per-channel Z-normalization (pre-normalization weighting). Table~\ref{tab:weighted_ablations} evaluates an alternative regime where these weights are explicitly applied post-normalization ($Z_j' = w_j \cdot Z_j$) to force the optimizer to recognize static priorities.

This exploratory single-run comparison shows that standard RVPO has the higher peak ($0.230$), while post-normalization weighted GDPO has the higher final score ($0.213$ versus $0.204$). The tested post-normalization RVPO settings score lower than standard RVPO in these runs. This 7B comparison of categorical weighting is distinct from the controlled 3B single-channel priority intervention in Appendix~\ref{app:corruption}.

\input{tab/weighted_ablation}

\subsection{Controlled Corruption and Priority Mismatch}
\label{app:corruption}

We target a fixed rubric position that is reproducibly the weakest channel in clean 3B RaR-Medicine runs. Each cell uses five seeds. We compare GDPO, RVPO ($k=0.5\to2.0$), and hard-min. The conditions are symmetric label noise at $p=0.25$ and $p=0.5$, a harsh judge with one-sided $1\to0$ flips at $p=0.25$, and a priority mismatch intervention that down-weights the weak channel to $0.2$ after normalization. Since we control the corruption, we record both the training signal and the underlying pre-corruption signal.

\input{tab/corruption_results}

Table~\ref{tab:corruption_results} reports target-channel aggregation weight normalized by its equal share $1/M$. GDPO has weight $1.0$ by construction in the unweighted reference and corruption conditions. Under symmetric noise at $p=0.5$, hard-min rises from $0.922$ to $2.095$ times equal share, while finite-$k$ RVPO rises from $1.009$ to $1.194$. The harsh-judge condition gives the same result. Neither formulation identifies the channel as unreliable; the finite-$k$ configuration concentrates less weight on it than hard-min in these corruption tests.

The difference between early-to-late drift in the observed and underlying signals remains within approximately $\pm0.015$ across arms and conditions. Corrupted-minus-reference HealthBench differences range from $-0.007$ to $+0.008$ across arms. We interpret this as no detectable downstream degradation at this sample size rather than proof of no effect.

Applying a post-normalization weight of $0.2$ reduces the target share to $0.826$ for RVPO and $0.304$ for hard-min. This offers a way to encode known priorities separately from difficulty. Positive multiplicative weights applied before per-channel normalization are absorbed by that normalization, whereas post-normalization weights change the aggregation inputs.

The tests are limited to 3B, and corruption is independent per generation. They probe unreliable judges rather than exploitability of judges whose errors correlate with an output feature.

\subsection{Explicit Variance Penalty (\texorpdfstring{$\beta$}{beta}) Ablation}
\label{app:beta_sweep}

Figure~\ref{fig:beta_sweep} shows the 7B Medicine sweep over constant values of the explicit variance coefficient $\beta$, with one training run per setting. Performance varies non-monotonically across the sweep, illustrating sensitivity to the penalty coefficient. Unlike SoftMin, the quadratic penalty grows without bound as inter-objective disagreement increases. Table~\ref{tab:rubrics_results} separately reports five-seed results for the selected $\beta=1.0$ setting.

\input{fig/beta_sweep}

\subsection{Broader Impacts}
\label{app:broader_impacts}

This work improves the ability of large language models to balance competing objectives and satisfy bottleneck constraints. By mitigating constraint neglect, RVPO provides a computationally efficient mechanism for encouraging more consistent behavior while maintaining competitive performance on the evaluated capability benchmarks. However, as a reward aggregation strategy, variance penalization is agnostic to the semantic nature of the constraints. While we apply it to enforce structural correctness and clinical accuracy, the same explicit constraint enforcement could theoretically be misused if malicious or biased reward models are intentionally provided to the optimizer.

\applefootnote{ \textcolor{textgray}{\sffamily Apple and the Apple logo are trademarks of Apple Inc., registered in the U.S. and other countries and regions.}}

\end{document}

%% file: apple_preamble.tex
\usepackage{amsmath}
\usepackage{enumerate}
\usepackage{algorithm}
\usepackage{algorithmic} 
\usepackage{amsfonts}
\usepackage{amsthm}
\usepackage{cleveref}
\usepackage{diagbox}
\usepackage{colortbl}
\usepackage{amssymb}
\usepackage{xspace}
\usepackage{wrapfig}
\usepackage{adjustbox}
\usepackage{tabularx}
\usepackage{booktabs}
\usepackage{mathtools}
\usepackage{tikz}
\usepackage{enumitem}
\usepackage{silence}
\usepackage{dsfont}
\usepackage[table]{xcolor}
\usepackage[dvipsnames]{xcolor}
\usepackage{multirow}
\usepackage{makecell}
\usepackage{xfakebold}
\input{math_commands}

\definecolor{textgray}{HTML}{6E6E73}
\usetikzlibrary{positioning, calc}
\usetikzlibrary{decorations.pathmorphing}

\makeatletter
\patchcmd{\wrong@fontshape}{\@gobbletwo}{}{}{}
\makeatother
\numberwithin{equation}{section}
\makeatletter
\AtBeginDocument{
  \urlstyle{sf}
  
}
\makeatother

\definecolor{light}{RGB}{125, 125, 125}
\crefname{tcb@cnt@pbox}{code}{code}
\Crefname{tcb@cnt@pbox}{Code}{Code}
\crefname{assumption}{assumption}{assumption}
\Crefname{assumption}{Assumption}{Assumptions}

\newtcolorbox[auto counter]{pbox}[2][]{
  colback=white,
  title=Code~\thetcbcounter: #2,
  #1,fonttitle=\sffamily,
  fontupper=\sffamily,
  arc=2pt,
  colframe=bgcolor,
  coltitle=fgcolor,
  colbacktitle=bgcolor,
  toptitle=0.25cm,
  bottomtitle=0.125cm
}

\makeatletter
\newcommand\applefootnote[1]{%
  \begingroup
  \renewcommand\thefootnote{}%
  \renewcommand\@makefntext[1]{\noindent##1}%
  \footnote{#1}%
  \addtocounter{footnote}{-1}%
  \endgroup
}
\makeatother

\definecolor{cverbbg}{gray}{0.90}

%% file: math_commands.tex
\usepackage{amsmath,amsfonts,bm}

\def\eqref#1{equation~\ref{#1}}

\def\1{\bm{1}}

\DeclareMathAlphabet{\mathsfit}{\encodingdefault}{\sfdefault}{m}{sl}
\SetMathAlphabet{\mathsfit}{bold}{\encodingdefault}{\sfdefault}{bx}{n}



%% file: fig/aggregation_gap.tex
\begin{figure}[htbp]
    \centering
    \resizebox{0.9\columnwidth}{!}{%
    \begin{tikzpicture}
        \begin{axis}[
            name=ax1,
            width=6.5cm, height=5.5cm,
            ymin=0, ymax=1.1,
            xmin=0.5, xmax=2.5,
            xtick={1, 2},
            xticklabels={Generation A, Generation B},
            ybar=0pt, 
            bar width=16pt,
            ylabel={\textbf{Aggregate Reward}},
            title={\textbf{Mean Aggregation (GRPO/GDPO)}},
            axis lines*=left,
            ymajorgrids=true,
            grid style={dotted, gray!40},
            legend to name=sharedlegend, 
            legend columns=-1,
            legend style={draw=none, font=\small, column sep=0.25em},
        ]
        \addplot[fill=blue!40!white, draw=black, thick] coordinates {(1, 1.0) (2, 0.5)};
        \addplot[fill=red!40!white, draw=black, thick] coordinates {(1, 0.03) (2, 0.5)};
        \legend{$R_1$ (e.g. Correctness), $R_2$ (e.g. Safety)}

        \draw[ultra thick, dashed, black] (axis cs:0.65, 0.5) -- (axis cs:1.35, 0.5);
        \draw[ultra thick, dashed, black] (axis cs:1.65, 0.5) -- (axis cs:2.35, 0.5);

        \node[anchor=south, inner sep=1.5pt] at (axis cs:1.27, 0.52) {\textbf{$A = 0.5$}};
        \node[anchor=south, inner sep=1.5pt] at (axis cs:2, 0.52) {\textbf{$A = 0.5$}};
        \end{axis}

        \begin{axis}[
            name=ax2,
            at={(ax1.south east)}, anchor=south west, xshift=2.0cm,
            width=6.5cm, height=5.5cm,
            ymin=0, ymax=1.1,
            xmin=0.5, xmax=2.5,
            xtick={1, 2},
            xticklabels={Generation A, Generation B},
            ybar=0pt,
            bar width=16pt,
            ylabel={\textbf{Aggregate Reward}}, 
            title={\textbf{Variance Regularization (RVPO)}},
            axis lines*=left,
            ymajorgrids=true,
            grid style={dotted, gray!40},
        ]
        \addplot[fill=blue!40!white, draw=black, thick] coordinates {(1, 1.0) (2, 0.5)};
        \addplot[fill=red!40!white, draw=black, thick] coordinates {(1, 0.03) (2, 0.5)};

        \draw[ultra thick, dashed, black] (axis cs:0.65, 0.08) -- (axis cs:1.35, 0.08);
        \draw[ultra thick, dashed, black] (axis cs:1.65, 0.5) -- (axis cs:2.35, 0.5);

        \draw[->, >=stealth, ultra thick, red!80!black] (axis cs:1.08, 0.56) -- (axis cs:1.08, 0.21);
        \node[anchor=west, align=left, text=red!80!black, font=\small\bfseries, inner sep=1pt] at (axis cs:1.1, 0.38) {Variance\\Penalty};
        
        \node[anchor=south, inner sep=1.5pt] at (axis cs:1.3, 0.1) {\textbf{$A \ll 0.5$}};
        \node[anchor=south, inner sep=1.5pt] at (axis cs:2, 0.52) {\textbf{$A = 0.5$}};
        \end{axis}
        
        \node[anchor=north, yshift=-0.1cm] at (current bounding box.south) {\ref*{sharedlegend}};
    \end{tikzpicture}%
    }
    \caption{\textbf{Constraint neglect in multi-objective RLHF.} \textbf{(Left)} Mean aggregation (GRPO/GDPO) treats outputs with critical constraint failures (Gen A) as mathematically identical to balanced outputs (Gen B). \textbf{(Right)} RVPO penalizes inter-reward variance, assigning a lower aggregate score to Gen A.}
    \label{fig:aggregation_gap}
\end{figure}

%% file: tab/rubrics_results_scale.tex
\begin{table}[t]
\centering
\caption{\textbf{Five-seed RaR-Medicine results on HealthBench.} Mean $\pm$ standard deviation over five distinct training seeds per method and scale, evaluated on all 5,000 questions using the micro-average of per-question rubric scores. Best is selected within each run; final is step 300. Bold marks the highest displayed mean in each size/checkpoint column, including ties.}
\label{tab:rubrics_results}
\begin{tabular}{llcc}
\toprule
\textbf{Size} & \textbf{Method} & \textbf{Best checkpoint} & \textbf{Final (300)} \\
\midrule
\multirow{6}{*}{3B} & GRPO (Implicit) & $0.167\pm0.008$ & $0.072\pm0.013$ \\
 & GRPO (Explicit) & $0.179\pm0.007$ & $0.064\pm0.043$ \\
 & GDPO & $0.178\pm0.011$ & $0.111\pm0.036$ \\
 & RVPO-min ($k=\infty$) & $\mathbf{0.180\pm0.008}$ & $0.136\pm0.012$ \\
 & RVPO-explicit ($\beta=1.0$) & $0.172\pm0.009$ & $0.023\pm0.025$ \\
 & RVPO ($k=0.5\to2.0$) & $\mathbf{0.180\pm0.005}$ & $\mathbf{0.154\pm0.020}$ \\
\midrule
\multirow{6}{*}{7B} & GRPO (Implicit) & $0.196\pm0.022$ & $\mathbf{0.160\pm0.036}$ \\
 & GRPO (Explicit) & $0.213\pm0.011$ & $0.035\pm0.044$ \\
 & GDPO & $0.192\pm0.006$ & $0.022\pm0.026$ \\
 & RVPO-min ($k=\infty$) & $0.190\pm0.013$ & $0.041\pm0.070$ \\
 & RVPO-explicit ($\beta=1.0$) & $\mathbf{0.215\pm0.007}$ & $\mathbf{0.160\pm0.035}$ \\
 & RVPO ($k=0.5\to2.0$) & $0.214\pm0.010$ & $0.152\pm0.040$ \\
\midrule
\multirow{6}{*}{14B} & GRPO (Implicit) & $0.242\pm0.011$ & $0.143\pm0.117$ \\
 & GRPO (Explicit) & $0.232\pm0.006$ & $0.039\pm0.078$ \\
 & GDPO & $0.207\pm0.032$ & $0.042\pm0.084$ \\
 & RVPO-min ($k=\infty$) & $0.243\pm0.005$ & $0.157\pm0.077$ \\
 & RVPO-explicit ($\beta=1.0$) & $0.222\pm0.018$ & $0.153\pm0.080$ \\
 & RVPO ($k=0.5\to2.0$) & $\mathbf{0.254\pm0.017}$ & $\mathbf{0.217\pm0.044}$ \\
\bottomrule
\end{tabular}
\end{table}

%% file: fig/healthbench_axes.tex
\begin{figure}[ht]
    \centering
    \includegraphics[width=\linewidth]{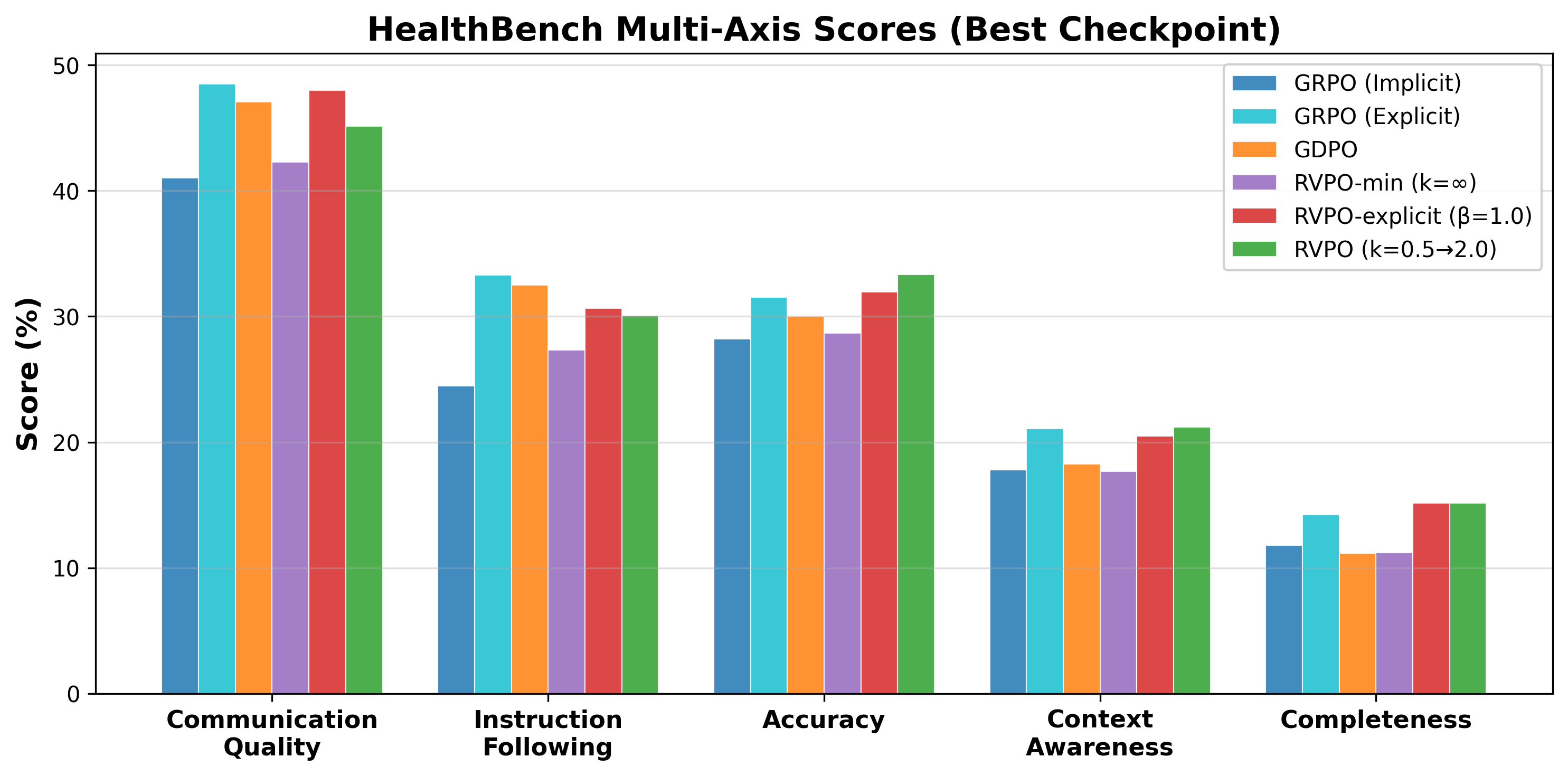}
    \caption{\textbf{Illustrative single-run per-axis comparison on HealthBench \citep{arora2025healthbench} (Medicine, Qwen2.5-7B).} Results at each method's best overall-HealthBench checkpoint, using the original single-run experiments. Five-seed overall results are reported in Table~\ref{tab:rubrics_results}.}
    \label{fig:healthbench_best}
\end{figure}

%% file: tab/bfclv3.tex
\begin{table}[ht]
\centering
\caption{\textbf{BFCL-v3 evaluation \citep{patil2025bfcl}.} Qwen2.5-1.5B and 3B. Avg is the mean of the three reported subcategories; AST Sum evaluates single and simple function calls; Parallel and Par.\ Multiple evaluate the ability to invoke multiple functions simultaneously. All methods achieve comparable accuracies on the evaluated tool-calling tasks. Results averaged across five runs per method (run-to-run std $\approx 0.4\%$).}
\label{tab:bfcl_results}
\vspace{0.5em}
\begin{tabular}{llcccc}
\toprule
\textbf{Size} & \textbf{Method} & \shortstack{\textbf{Avg}\\\textbf{(\%)}} & \shortstack{\textbf{AST Sum}\\\textbf{(\%)}} & \shortstack{\textbf{Parallel}\\\textbf{(\%)}} & \shortstack{\textbf{Par. Multiple}\\\textbf{(\%)}} \\
\midrule
\multirow{4}{*}{1.5B}
 & GRPO & 70.4 & 71.7 & 70.3 & 69.2 \\
 & GDPO & \textbf{71.9} & \textbf{73.4} & 71.7 & \textbf{70.7} \\
 & RVPO-explicit ($\beta=1.0$) & 71.3 & 72.5 & 71.0 & 70.5 \\
 & RVPO ($k=1.0$) & \textbf{71.9} & 72.9 & \textbf{72.1} & 70.6 \\
\midrule
\multirow{4}{*}{3B}
 & GRPO & 80.3 & 81.3 & 77.8 & 81.9 \\
 & GDPO & 80.7 & 81.6 & 78.8 & 81.7 \\
 & RVPO-explicit ($\beta=1.0$) & \textbf{81.4} & \textbf{82.1} & \textbf{80.1} & \textbf{81.9} \\
 & RVPO ($k=1.0$) & 80.4 & 81.5 & 78.4 & 81.3 \\
\bottomrule
\end{tabular}
\end{table}

%% file: fig/tool_plots.tex
\begin{figure}[ht]
    \centering
    
    
    \includegraphics[width=0.49\textwidth]{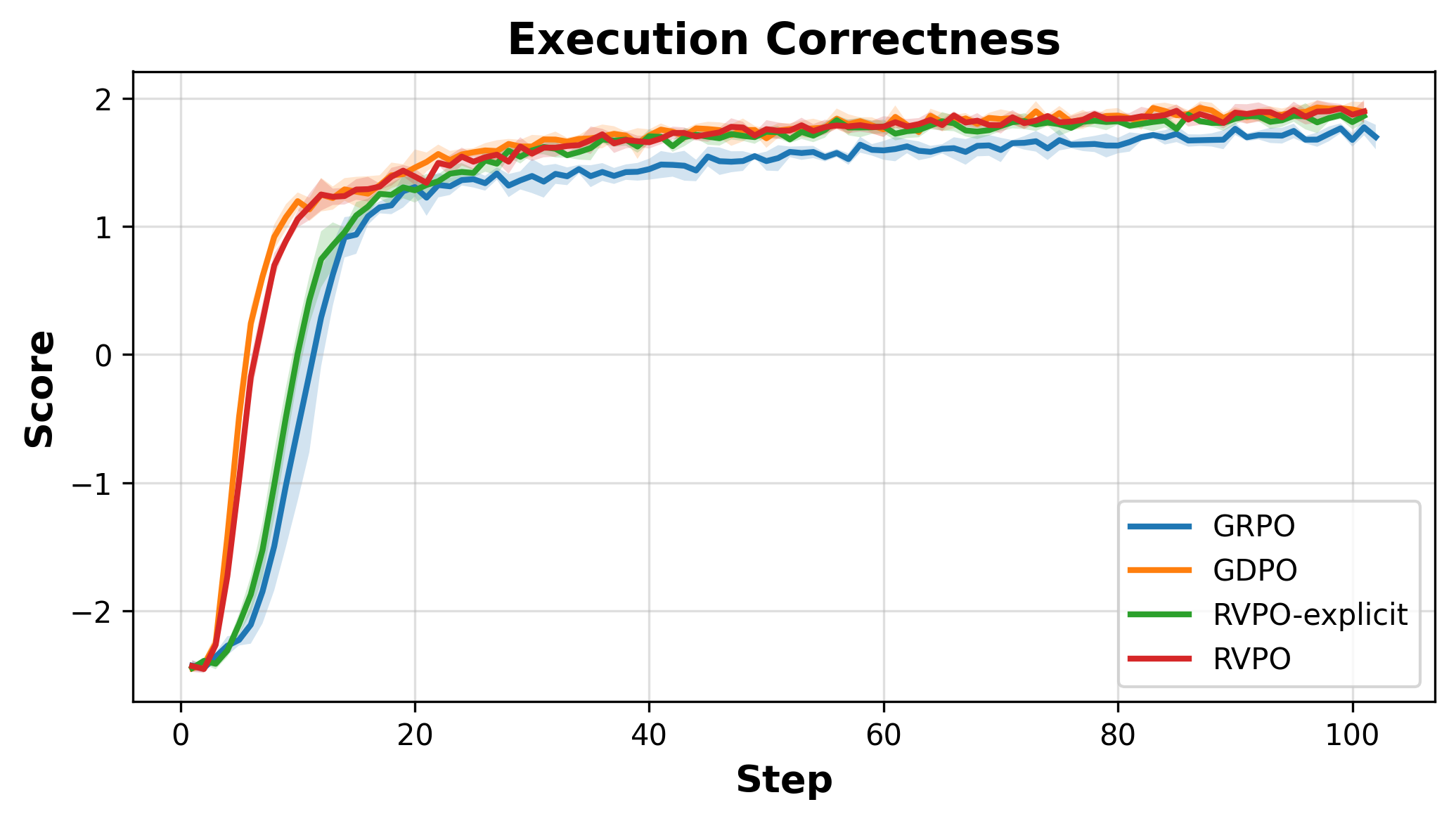}\hfill
    \includegraphics[width=0.49\textwidth]{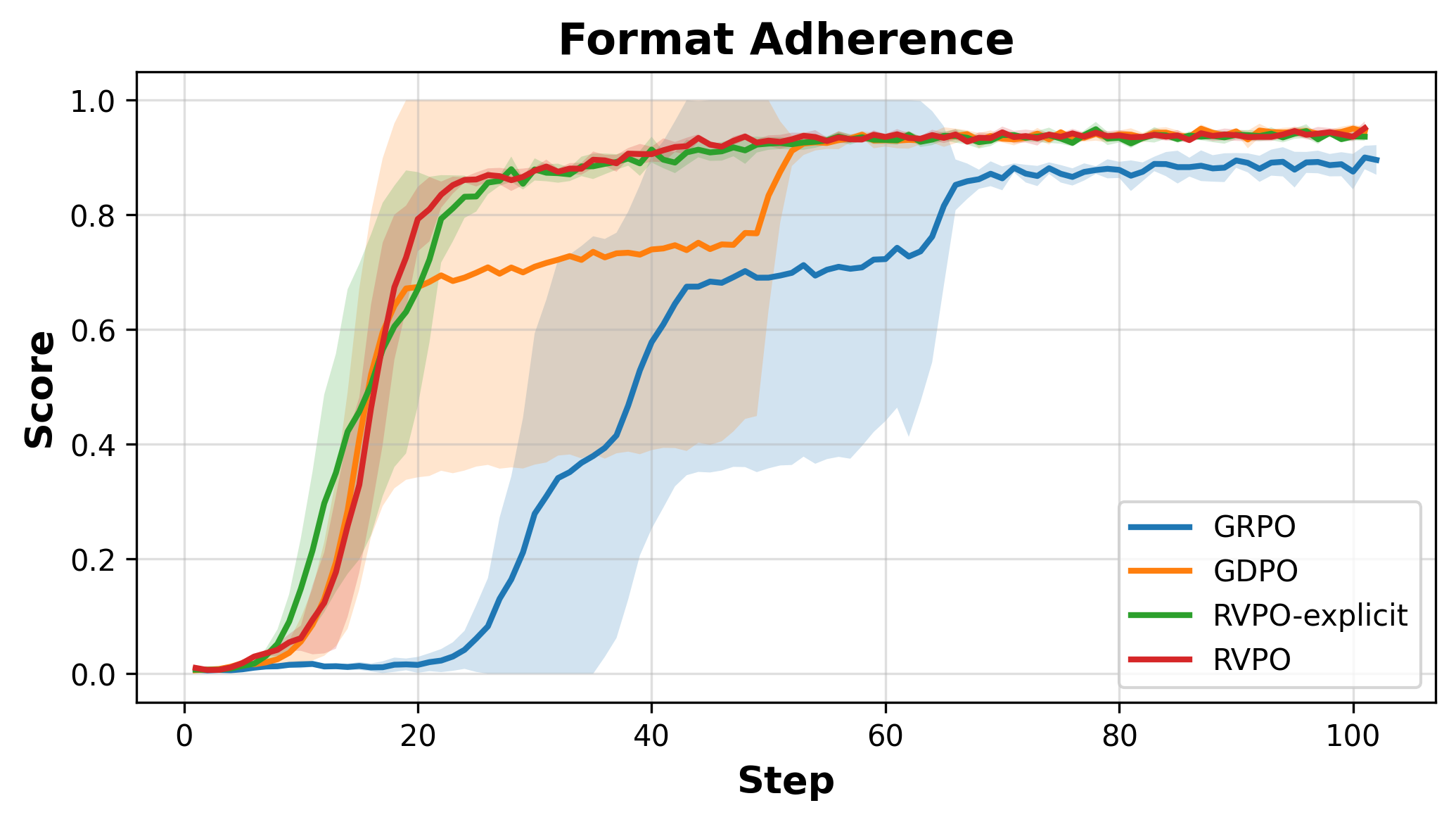}
    
    \caption{\textbf{Tool calling (RLLA) training dynamics \citep{qian2025toolrl}.} Qwen2.5-1.5B training progression across five independent runs; solid lines show the mean and shaded regions $\pm 1$ standard deviation. \textit{(Left)} Execution correctness. \textit{(Right)} Format adherence. RVPO and RVPO-explicit improve format adherence while maintaining execution correctness.
    }
    \label{fig:tool_calling_results}
\end{figure}

%% file: tab/science_results.tex
\begin{table}[ht]
\centering
\caption{\textbf{RaR-Science GPQA-Diamond evaluation.} Loglikelihood accuracy on 198 questions, one training run per configuration. Best and final checkpoints are reported as a descriptive parity check.}
\label{tab:science_results}
\begin{tabular}{llcc}
\toprule
\textbf{Size} & \textbf{Method} & \textbf{Best checkpoint} & \textbf{Final (300)} \\
\midrule
\multirow{6}{*}{3B} & GRPO (Implicit) & 0.338 & 0.313 \\
 & GRPO (Explicit) & 0.343 & 0.343 \\
 & GDPO & 0.308 & 0.293 \\
 & RVPO-min ($k=\infty$) & 0.348 & 0.283 \\
 & RVPO-explicit ($\beta=0.5$) & 0.313 & 0.308 \\
 & RVPO ($k=1.0\to2.0$) & 0.313 & 0.303 \\
\midrule
\multirow{6}{*}{7B} & GRPO (Implicit) & 0.318 & 0.283 \\
 & GRPO (Explicit) & 0.343 & 0.343 \\
 & GDPO & 0.323 & 0.318 \\
 & RVPO-min ($k=\infty$) & 0.323 & 0.308 \\
 & RVPO-explicit ($\beta=0.5$) & 0.318 & 0.318 \\
 & RVPO ($k=1.0\to2.0$) & 0.338 & 0.338 \\
\midrule
\multirow{6}{*}{14B} & GRPO (Implicit) & 0.394 & 0.359 \\
 & GRPO (Explicit) & 0.404 & 0.338 \\
 & GDPO & 0.394 & 0.374 \\
 & RVPO-min ($k=\infty$) & 0.369 & 0.293 \\
 & RVPO-explicit ($\beta=0.5$) & 0.414 & 0.359 \\
 & RVPO ($k=1.0\to2.0$) & 0.384 & 0.384 \\
\bottomrule
\end{tabular}
\end{table}

%% file: tab/k_sweep_ablation.tex
\begin{table}[ht]
\centering
\caption{\textbf{Risk coefficient selection on HealthBench (Medicine, Qwen2.5-7B).} Original exploratory results, one training run per configuration. The selected schedule favors final-checkpoint retention over the highest early peak; repeated-run method comparisons are reported separately in Table~\ref{tab:rubrics_results}.}
\label{tab:k_sweep_appendix}
\vspace{0.5em}
\begin{tabular}{lcc}
\toprule
\textbf{Schedule ($k$)} & \textbf{Best Ckpt} & \textbf{Final (300)} \\
\midrule
\multicolumn{3}{l}{\textit{Baselines}} \\
GRPO (Implicit) & 0.193 & 0.193 \\
GRPO (Explicit) & 0.221 & 0.102 \\
GDPO ($k=0$) & 0.198 & 0.026 \\
RVPO-min ($k=\infty$) & 0.191 & 0.000 \\
\midrule
\multicolumn{3}{l}{\textit{Constant Schedules}} \\
$k=0.25$ & 0.207 & 0.129 \\
$k=0.5$ & 0.171 & 0.095 \\
$k=1.0$ & 0.194 & 0.015 \\
$k=2.0$ & 0.225 & 0.083 \\
$k=5.0$ & \textbf{0.241} & 0.095 \\
$k=10.0$ & 0.202 & 0.010 \\
\midrule
\multicolumn{3}{l}{\textit{Annealed Curricula}} \\
$0.1 \to 2.0$ & 0.214 & 0.132 \\
$0.25 \to 1.0$ & 0.209 & 0.082 \\
$0.5 \to 5.0$ & 0.198 & 0.139 \\
\textbf{$0.5 \to 2.0$} & 0.230 & \textbf{0.204} \\
$1.0 \to 2.0$ & 0.143 & 0.000 \\
\bottomrule
\end{tabular}
\end{table}

%% file: tab/weighted_ablation.tex
\begin{table}[h]
\centering
\caption{\textbf{Pre- vs.\ post-normalization weighting on HealthBench (Medicine, Qwen2.5-7B).} Original exploratory results, one training run per configuration. Standard RVPO has the highest peak; post-normalization weighted GDPO has the highest final score. GRPO-explicit is the raw weighted-scalar baseline.}
\label{tab:weighted_ablations}
\vspace{0.5em}
\begin{tabular}{lcc}
\toprule
\textbf{Method} & \textbf{Best Checkpoint} & \textbf{Final (Step 300)} \\
\midrule
\multicolumn{3}{l}{\textit{Pre-normalization Weighting (Standard)}} \\
GDPO (Mean) & 0.198 & 0.026 \\
RVPO ($k=0.5 \to 2.0$) & \textbf{0.230} & 0.204 \\
\midrule
\multicolumn{3}{l}{\textit{Weighted Baselines (GRPO raw; GDPO post-normalization)}} \\
GRPO (Explicit) & 0.221 & 0.102 \\
GDPO & 0.213 & \textbf{0.213} \\
\midrule
\multicolumn{3}{l}{\textit{Post-normalization Weighting (RVPO)}} \\
RVPO ($k=0.5 \to 2.0$) & 0.192 & 0.078 \\
RVPO ($k=1.0 \to 2.0$) & 0.201 & 0.000 \\
RVPO ($k=2.0$ const) & 0.160 & 0.106 \\
\bottomrule
\end{tabular}
\end{table}

%% file: tab/corruption_results.tex
\begin{table}[ht]
\centering
\caption{\textbf{Controlled corruption and priority intervention at 3B.} Target-channel aggregation weight divided by equal share $1/M$ (five seeds per cell, mean $\pm$ standard deviation). The priority intervention is distinct from label corruption.}
\label{tab:corruption_results}
\begin{tabular}{lcc}
\toprule
\textbf{Condition} & \textbf{RVPO ($k=0.5\to2$)} & \textbf{Hard-min} \\
\midrule
Reference ($p=0$) & $1.009 \pm 0.003$ & $0.922 \pm 0.021$ \\
Symmetric noise ($p=0.25$) & $1.122 \pm 0.008$ & $1.703 \pm 0.034$ \\
Symmetric noise ($p=0.5$) & $1.194 \pm 0.010$ & $2.095 \pm 0.042$ \\
Harsh judge ($p=0.25$) & $1.060 \pm 0.010$ & $1.174 \pm 0.024$ \\
Priority weight $0.2$ & $0.826 \pm 0.003$ & $0.304 \pm 0.011$ \\
\bottomrule
\end{tabular}
\end{table}

%% file: fig/beta_sweep.tex
\begin{figure}[ht]
    \centering
    \includegraphics[width=0.75\linewidth]{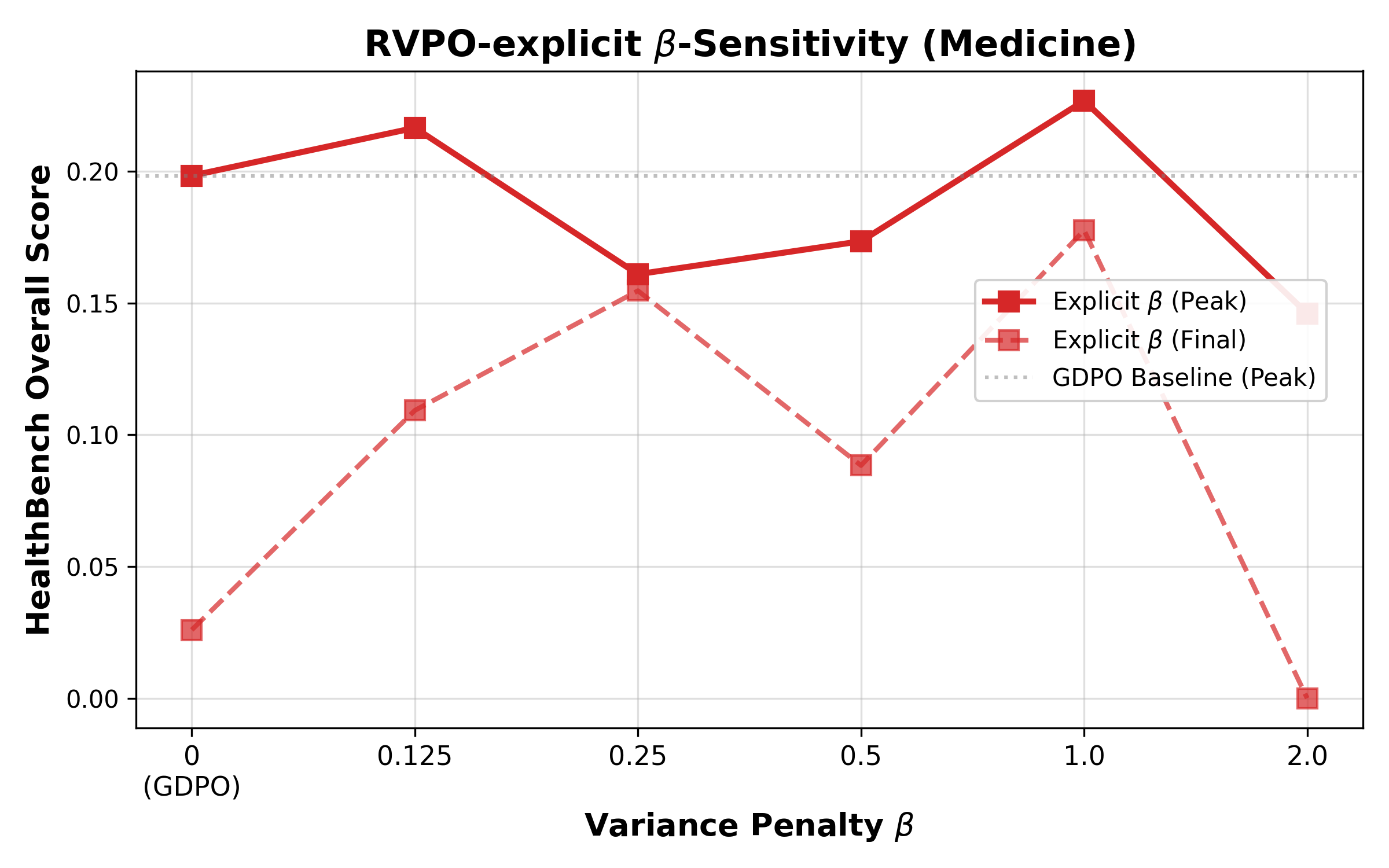}
    \caption{\textbf{Explicit variance penalty ($\beta$) sweep on HealthBench (Medicine, Qwen2.5-7B).} Best- and final-checkpoint scores for constant penalty coefficients, with one training run per setting. The selected $\beta=1.0$ configuration is evaluated across five seeds in Table~\ref{tab:rubrics_results}.}
    \label{fig:beta_sweep}
\end{figure}

%% file: references.bib
@article{ouyang2022training,
  title={Training language models to follow instructions with human feedback},
  author={Ouyang, Long and Wu, Jeffrey and Jiang, Xu and Almeida, Diogo and Wainwright, Carroll and Mishkin, Pamela and Zhang, Chong and Agarwal, Sandhini and Slama, Katarina and Ray, Alex and others},
  journal={Advances in neural information processing systems},
  volume={35},
  pages={27730--27744},
  year={2022}
}

@article{rafailov2023direct,
  title={Direct preference optimization: Your language model is secretly a reward model},
  author={Rafailov, Rafael and Sharma, Archit and Mitchell, Eric and Manning, Christopher D and Ermon, Stefano and Finn, Chelsea},
  journal={Advances in neural information processing systems},
  volume={36},
  pages={53728--53741},
  year={2023}
}

@inproceedings{chen2024accuracy,
  title={The accuracy paradox in {RLHF}: When better reward models don’t yield better language models},
  author={Chen, Yanjun and Zhu, Dawei and Sun, Yirong and Chen, Xinghao and Zhang, Wei and Shen, Xiaoyu},
  booktitle={Proceedings of the 2024 Conference on Empirical Methods in Natural Language Processing},
  pages={2980--2989},
  year={2024}
}

@article{dai2025mitigating,
  title={Mitigating reward over-optimization in {RLHF} via behavior-supported regularization},
  author={Dai, Juntao and Chen, Taiye and Yang, Yaodong and Zheng, Qian and Pan, Gang},
  journal={arXiv preprint arXiv:2503.18130},
  year={2025}
}

@inproceedings{gao2023scaling,
  title={Scaling laws for reward model overoptimization},
  author={Gao, Leo and Schulman, John and Hilton, Jacob},
  booktitle={International Conference on Machine Learning},
  pages={10835--10866},
  year={2023},
  organization={PMLR}
}

@article{li2025optimizing,
  title={Optimizing safe and aligned language generation: A multi-objective {GRPO} approach},
  author={Li, Xuying and Li, Zhuo and Kosuga, Yuji and Bian, Victor},
  journal={arXiv preprint arXiv:2503.21819},
  year={2025}
}

@article{hayes2022practical,
  title={A practical guide to multi-objective reinforcement learning and planning},
  author={Hayes, Conor F and R{\u{a}}dulescu, Roxana and Bargiacchi, Eugenio and K{\"a}llstr{\"o}m, Johan and Macfarlane, Matthew and Reymond, Mathieu and Verstraeten, Timothy and Zintgraf, Luisa M and Dazeley, Richard and Heintz, Fredrik and others},
  journal={Autonomous Agents and Multi-Agent Systems},
  volume={36},
  number={1},
  pages={26},
  year={2022},
  publisher={Springer}
}

@article{kim2025beyond,
  title={Beyond {RLHF} and {NLHF}: Population-Proportional Alignment under an Axiomatic Framework},
  author={Kim, Kihyun and Zhang, Jiawei and Ozdaglar, Asuman and Parrilo, Pablo A},
  journal={arXiv preprint arXiv:2506.05619},
  year={2025}
}

@article{shao2024deepseekmath,
  title={{DeepSeekMath}: Pushing the limits of mathematical reasoning in open language models},
  author={Shao, Zhihong and Wang, Peiyi and Zhu, Qihao and Xu, Runxin and Song, Junxiao and Bi, Xiao and Zhang, Haowei and Zhang, Mingchuan and Li, YK and others},
  journal={arXiv preprint arXiv:2402.03300},
  year={2024}
}

@article{liu2026gdpo,
  title={{GDPO}: Group reward-decoupled normalization policy optimization for multi-reward {RL} optimization},
  author={Liu, Shih-Yang and Dong, Xin and Lu, Ximing and Diao, Shizhe and Belcak, Peter and Liu, Mingjie and Chen, Min-Hung and Yin, Hongxu and Wang, Yu-Chiang Frank and Cheng, Kwang-Ting and others},
  journal={arXiv preprint arXiv:2601.05242},
  year={2026}
}

@article{howard1972risk,
  title={Risk-sensitive Markov decision processes},
  author={Howard, Ronald A and Matheson, James E},
  journal={Management science},
  volume={18},
  number={7},
  pages={356--369},
  year={1972},
  publisher={INFORMS}
}

@inproceedings{hu2023tighter,
  title={A tighter problem-dependent regret bound for risk-sensitive reinforcement learning},
  author={Hu, Xiaoyan and Leung, Ho-fung},
  booktitle={International Conference on Artificial Intelligence and Statistics},
  pages={5411--5437},
  year={2023},
  organization={PMLR}
}

@article{zhong2025risk,
  title={Risk-sensitive deep {RL}: Variance-constrained actor-critic provably finds globally optimal policy},
  author={Zhong, Han and Deng, Xun and Fang, Ethan X and Yang, Zhuoran and Wang, Zhaoran and Li, Runze},
  journal={Journal of the American Statistical Association},
  pages={1077--1089},
  year={2026},
  doi={10.1080/01621459.2025.2583501},
  publisher={Taylor \& Francis}
}

@article{qian2025toolrl,
  title={{ToolRL}: Reward is all tool learning needs},
  author={Qian, Cheng and Acikgoz, Emre Can and He, Qi and Wang, Hongru and Chen, Xiusi and Hakkani-T{\"u}r, Dilek and Tur, Gokhan and Ji, Heng},
  journal={arXiv preprint arXiv:2504.13958},
  year={2025}
}

@inproceedings{patil2025bfcl,
  title={The {Berkeley Function Calling Leaderboard} ({BFCL}): From Tool Use to Agentic Evaluation of Large Language Models}, 
author={Patil, Shishir G. and Mao, Huanzhi and Yan, Fanjia and Ji, Charlie Cheng-Jie and Suresh, Vishnu and Stoica, Ion and Gonzalez, Joseph E.},
booktitle={Forty-second International Conference on Machine Learning},
year={2025},
}

@article{gunjal2025rubrics,
  title={Rubrics as rewards: Reinforcement learning beyond verifiable domains},
  author={Gunjal, Anisha and Wang, Anthony and Lau, Elaine and Nath, Vaskar and He, Yunzhong and Liu, Bing and Hendryx, Sean},
  journal={arXiv preprint arXiv:2507.17746},
  year={2025}
}

@article{meng2024simpo,
  title={{SimPO}: Simple preference optimization with a reference-free reward},
  author={Meng, Yu and Xia, Mengzhou and Chen, Danqi},
  journal={Advances in Neural Information Processing Systems},
  volume={37},
  pages={124198--124235},
  year={2024}
}

@inproceedings{park2024disentangling,
  title={Disentangling length from quality in direct preference optimization},
  author={Park, Ryan and Rafailov, Rafael and Ermon, Stefano and Finn, Chelsea},
  booktitle={Findings of the Association for Computational Linguistics: ACL 2024},
  pages={4998--5017},
  year={2024}
}

@article{chen2024odin,
  title={{ODIN}: Disentangled reward mitigates hacking in {RLHF}},
  author={Chen, Lichang and Zhu, Chen and Soselia, Davit and Chen, Jiuhai and Zhou, Tianyi and Goldstein, Tom and Huang, Heng and Shoeybi, Mohammad and Catanzaro, Bryan},
  journal={arXiv preprint arXiv:2402.07319},
  year={2024}
}

@inproceedings{achiam2017constrained,
  title={Constrained policy optimization},
  author={Achiam, Joshua and Held, David and Tamar, Aviv and Abbeel, Pieter},
  booktitle={International conference on machine learning},
  pages={22--31},
  year={2017},
  organization={Pmlr}
}

@article{dai2023safe,
  title={Safe {RLHF}: Safe reinforcement learning from human feedback},
  author={Dai, Josef and Pan, Xuehai and Sun, Ruiyang and Ji, Jiaming and Xu, Xinbo and Liu, Mickel and Wang, Yizhou and Yang, Yaodong},
  journal={arXiv preprint arXiv:2310.12773},
  year={2023}
}

@article{rame2023rewarded,
  title={Rewarded soups: towards pareto-optimal alignment by interpolating weights fine-tuned on diverse rewards},
  author={Rame, Alexandre and Couairon, Guillaume and Dancette, Corentin and Gaya, Jean-Baptiste and Shukor, Mustafa and Soulier, Laure and Cord, Matthieu},
  journal={Advances in Neural Information Processing Systems},
  volume={36},
  pages={71095--71134},
  year={2023}
}

@article{viswanathan2025checklists,
  title={Checklists are better than reward models for aligning language models},
  author={Viswanathan, Vijay and Sun, Yanchao and Ma, Shuang and Kong, Xiang and Cao, Meng and Neubig, Graham and Wu, Tongshuang},
  journal={arXiv preprint arXiv:2507.18624},
  year={2025}
}

@article{mu2024rule,
  title={Rule based rewards for language model safety},
  author={Mu, Tong and Helyar, Alec and Heidecke, Johannes and Achiam, Joshua and Vallone, Andrea and Kivlichan, Ian and Lin, Molly and Beutel, Alex and Schulman, John and Weng, Lilian},
  journal={Advances in Neural Information Processing Systems},
  volume={37},
  pages={108877--108901},
  year={2024}
}

@inproceedings{hashemi2024llm,
  title={{LLM}-rubric: A multidimensional, calibrated approach to automated evaluation of natural language texts},
  author={Hashemi, Helia and Eisner, Jason and Rosset, Corby and Van Durme, Benjamin and Kedzie, Chris},
  booktitle={Proceedings of the 62nd Annual Meeting of the Association for Computational Linguistics (Volume 1: Long Papers)},
  pages={13806--13834},
  year={2024}
}

@inproceedings{ahmadian2024back,
  title={Back to basics: Revisiting REINFORCE-style optimization for learning from human feedback in LLMs},
  author={Ahmadian, Arash and Cremer, Chris and Gall{\'e}, Matthias and Fadaee, Marzieh and Kreutzer, Julia and Pietquin, Olivier and {\"U}st{\"u}n, Ahmet and Hooker, Sara},
  booktitle={Proceedings of the 62nd Annual Meeting of the Association for Computational Linguistics (Volume 1: Long Papers)},
  pages={12248--12267},
  year={2024}
}

@article{guo2025deepseek,
  title={{DeepSeek-R1}: Incentivizing reasoning capability in {LLMs} via reinforcement learning},
  author={Guo, Daya and Yang, Dejian and Zhang, Haowei and Song, Junxiao and Wang, Peiyi and Zhu, Qihao and Xu, Runxin and Zhang, Ruoyu and Ma, Shirong and Bi, Xiao and others},
  journal={arXiv preprint arXiv:2501.12948},
  year={2025}
}

@article{yu2025dapo,
  title={{DAPO}: An open-source {LLM} reinforcement learning system at scale},
  author={Yu, Qiying and Zhang, Zheng and Zhu, Ruofei and Yuan, Yufeng and Zuo, Xiaochen and Yue, Yu and Dai, Weinan and Fan, Tiantian and Liu, Gaohong and Liu, Lingjun and others},
  journal={arXiv preprint arXiv:2503.14476},
  year={2025}
}

@inproceedings{asadi2017alternative,
  title={An alternative softmax operator for reinforcement learning},
  author={Asadi, Kavosh and Littman, Michael L},
  booktitle={International Conference on Machine Learning},
  pages={243--252},
  year={2017},
  organization={PMLR}
}

@article{paternain2019constrained,
  title={Constrained reinforcement learning has zero duality gap},
  author={Paternain, Santiago and Chamon, Luiz and Calvo-Fullana, Miguel and Ribeiro, Alejandro},
  journal={Advances in Neural Information Processing Systems},
  volume={32},
  year={2019}
}

@article{arora2025healthbench,
  title={{HealthBench}: Evaluating large language models towards improved human health},
  author={Arora, Rahul K and Wei, Jason and Hicks, Rebecca Soskin and Bowman, Preston and Qui{\~n}onero-Candela, Joaquin and Tsimpourlas, Foivos and Sharman, Michael and Shah, Meghan and Vallone, Andrea and Beutel, Alex and others},
  journal={arXiv preprint arXiv:2505.08775},
  year={2025}
}

@inproceedings{rein2024gpqa,
  title={{GPQA}: A graduate-level {Google}-proof {Q\&A} benchmark},
  author={Rein, David and Hou, Betty Li and Stickland, Asa Cooper and Petty, Jackson and Pang, Richard Yuanzhe and Dirani, Julien and Michael, Julian and Bowman, Samuel R},
  booktitle={First conference on language modeling},
  year={2024}
}

@article{yang2024qwen2,
  title={Qwen2.5 Technical Report},
  author={Yang, An and Yang, Baosong and Zhang, Beichen and Hui, Binyuan and Zheng, Bo and Yu, Bowen and Li, Chengyuan and Liu, Dayiheng and Huang, Fei and Wei, Haoran and others},
  journal={arXiv preprint arXiv:2412.15115},
  year={2024}
}

@article{sheng2024hybridflow,
  title   = {HybridFlow: A Flexible and Efficient RLHF Framework},
  author  = {Guangming Sheng and Chi Zhang and Zilingfeng Ye and Xibin Wu and Wang Zhang and Ru Zhang and Yanghua Peng and Haibin Lin and Chuan Wu},
  year    = {2024},
  journal = {arXiv preprint arXiv: 2409.19256}
}

@inproceedings{nguyen2025multi,
  title={Multi-attribute steering of language models via targeted intervention},
  author={Nguyen, Duy and Prasad, Archiki and Stengel-Eskin, Elias and Bansal, Mohit},
  booktitle={Proceedings of the 63rd Annual Meeting of the Association for Computational Linguistics (Volume 1: Long Papers)},
  pages={20619--20634},
  year={2025}
}

@article{lin2025parm,
  title={{PARM}: Multi-objective test-time alignment via preference-aware autoregressive reward model},
  author={Lin, Baijiong and Jiang, Weisen and Xu, Yuancheng and Chen, Hao and Chen, Ying-Cong},
  journal={arXiv preprint arXiv:2505.06274},
  year={2025}
}

@article{wang2024helpsteer,
  title={{HelpSteer} 2: Open-source dataset for training top-performing reward models},
  author={Wang, Zhilin and Dong, Yi and Delalleau, Olivier and Zeng, Jiaqi and Shen, Gerald and Egert, Daniel and Zhang, Jimmy J and Sreedhar, Makesh N and Kuchaiev, Oleksii},
  journal={Advances in Neural Information Processing Systems},
  volume={37},
  pages={1474--1501},
  year={2024}
}

@article{chen2026reward,
  title={Reward-free Alignment for Conflicting Objectives},
  author={Chen, Peter L and Li, Xiaopeng and Chen, Xi and Lin, Tianyi},
  journal={arXiv preprint arXiv:2602.02495},
  year={2026}
}

@article{atkinson1970measurement,
  title={On the Measurement of Inequality},
  author={Atkinson, Anthony B.},
  journal={Journal of Economic Theory},
  volume={2},
  number={3},
  pages={244--263},
  year={1970},
  doi={10.1016/0022-0531(70)90039-6}
}

@article{roijers2013survey,
  title={A Survey of Multi-Objective Sequential Decision-Making},
  author={Roijers, Diederik M. and Vamplew, Peter and Whiteson, Shimon and Dazeley, Richard},
  journal={Journal of Artificial Intelligence Research},
  volume={48},
  pages={67--113},
  year={2013},
  doi={10.1613/jair.3987}
}

@inproceedings{chakraborty2024maxmin,
  title={{MaxMin-RLHF}: Alignment with Diverse Human Preferences},
  author={Chakraborty, Souradip and Qiu, Jiahao and Yuan, Hui and Koppel, Alec and Manocha, Dinesh and Huang, Furong and Bedi, Amrit and Wang, Mengdi},
  booktitle={Proceedings of the 41st International Conference on Machine Learning},
  series={Proceedings of Machine Learning Research},
  volume={235},
  pages={6116--6135},
  year={2024},
  publisher={PMLR},
  url={https://proceedings.mlr.press/v235/chakraborty24b.html}
}
